\documentclass[letter, 10 pt, conference]{ieeeconf}  

\IEEEoverridecommandlockouts 
\usepackage[english]{babel}
\usepackage[utf8]{inputenc}
\usepackage{times} 
\usepackage{cite}

\usepackage{booktabs}

\usepackage[hidelinks]{hyperref}
\usepackage[nolist]{acronym}

\usepackage{multirow}

\usepackage{subcaption}

\usepackage{siunitx}
\usepackage{graphicx} 
\usepackage{epsfig} 
\usepackage{pgfplots}
\pgfplotsset{compat=1.18}
\usepgfplotslibrary{statistics}
\usepgfplotslibrary{groupplots}

\usepackage{float}

\usepackage[export]{adjustbox}

\usepackage{xcolor}
\definecolor{myYellow}{rgb}{0.93,0.69,0.13}
\definecolor{myPurple}{rgb}{0.49,0.18,0.56}
\definecolor{myGreen}{rgb}{0.26 0.72 0.54}
\definecolor{darkgreen}{rgb}{0.272, 0.50, 0.376}
\definecolor{lightgreen}{rgb}{0.585, 0.82, 0.647}

\colorlet{mydarkblue}{blue!30!black}

\usepackage{mathtools}
\usepackage{nicefrac}
\usepackage{amsmath}
\usepackage{amssymb}
\usepackage{bm} 
\usepackage{scalerel} 

\DeclareMathAlphabet{\pazocal}{OMS}{zplm}{m}{n}

\DeclareMathOperator*{\minimize}{minimize}

\usepackage{etoolbox}
\makeatletter%
\AfterPreamble{%
	\usepackage{hyperref}%
	\let\oldhypertarget\hypertarget%
	\renewcommand{\hypertarget}[2]{%
		\oldhypertarget{#1}{#2}%
		\protected@write\@mainaux{}{%
			\string\expandafter\string\gdef%
			\string\csname\string\detokenize{#1}\string\endcsname{#2}%
		}%
	}%
	\newcommand{\myhyperlink}[1]{%
		\hyperlink{#1}{\csname #1\endcsname}%
	}%
}
\makeatother%

\newcounter{Remark}
\newcounter{Problem}
\usepackage{algorithm}
\usepackage[noend]{algpseudocode}

\makeatletter
\def\BState{\State\hskip-\ALG@thistlm}
\makeatother

\usepackage{tikz}
\pgfdeclarelayer{foreground}
\pgfsetlayers{background, main, foreground}
\usetikzlibrary{quotes, angles, backgrounds, arrows, automata, shapes, positioning, calc, through, spy, decorations.pathreplacing, decorations.markings, arrows.meta, automata, petri, shapes.multipart}

\tikzset{
    imglabel/.style={
      rectangle,
      inner sep=2pt,
      text=black,
      minimum height=1em,
      text centered,
      fill=white,
      fill opacity=1.0,
      text opacity=1,
      anchor=south west,
    },
  }
\tikzset{
	state/.style={
		rectangle,
		draw=black, very thick,
		minimum height=1.0em,
		text centered,
	},
}
\tikzset{
  on each segment/.style={
    decorate,
    decoration={
      show path construction,
      moveto code={},
      lineto code={
        \path [#1]
        (\tikzinputsegmentfirst) -- (\tikzinputsegmentlast);
      },
      curveto code={
        \path [#1] (\tikzinputsegmentfirst)
        .. controls
        (\tikzinputsegmentsupporta) and (\tikzinputsegmentsupportb)
        ..
        (\tikzinputsegmentlast);
      },
      closepath code={
        \path [#1]
        (\tikzinputsegmentfirst) -- (\tikzinputsegmentlast);
      },
    },
  },
  mid arrow/.style={postaction={decorate,decoration={
        markings,
        mark=at position .5 with {\arrow[#1]{stealth}}
      }}},
}

\tikzset{
  half circle/.style={
      semicircle,
      shape border rotate=180,
      anchor=chord center,
      minimum size=5mm
      }
}

\graphicspath{{./figures/}}

\newcommand\copyrighttext{%
    \small \begin{center} \color{red} \textcopyright\,2026 IEEE. Accepted for presentation to the ``2026 International Conference on Unmanned Aircraft Systems (ICUAS)", 15–18 June 2026, Corfu, Greece. Personal use of this material is permitted. Permission from IEEE must be obtained for all other uses, in any current or future media, including reprinting/republishing this material for advertising or promotional purposes, creating new collective works, for resale or redistribution to servers or lists, or reuse of any copyrighted component of this work in other works. \end{center}}
\newcommand\copyrightnotice{%
	\begin{tikzpicture}[remember picture,overlay]
	\node[anchor=south,yshift=25.6cm] at (current page.south) 
	{\color{red}\fbox{\parbox{\dimexpr\textwidth-\fboxsep-\fboxrule\relax}{\copyrighttext}}};
	\end{tikzpicture}%
}

\title{\copyrightnotice \LARGE \bf Sensitivity-Based Tube NMPC for Cooperative Aerial Structures Under Parametric Uncertainty} 

\author{Giuseppe Silano$^{1}$, Quentin Sablé$^{2}$, Marco Tognon$^{3}$, Luigi Iannelli$^{4}$, and Antonio Franchi$^{2,5}$  
    \thanks{This work was partially funded by the research fund for the Italian Electrical System (decree n. 388, Nov.~6th, 2024), the EU's HE AUTOASSESS project no.~101120732, the GAČR project no.~26-22419S, and ANR-23-CE33-0013 ``FlyHandyBot''.
    $^1$Ricerca sul Sistema Energetico S.p.A., Milan, Italy, and Czech Technical University, Prague, Czechia (e-mail: {\tt\small silangiu@fel.cvut.cz}). $^2$University of Twente, Enschede, The Netherlands (emails: {\tt\footnotesize q.l.g.sable@utwente.nl, schol@r-franchi.eu}). $^3$University of Rennes, CNRS, Inria, IRISA, 35042, France (e-mail: {\tt\small marco.tognon@inria.fr)}. $^4$University of Sannio, Benevento, Italy (e-mail: {\tt\small luiannel@unisannio.it)}. $^5$Sapienza University of Rome, Rome, Italy.} 
}

\begin{document}

\maketitle
\thispagestyle{empty} 
\pagestyle{empty} 

\setlength{\textfloatsep}{6pt plus 2pt minus 2pt}
\setlength{\floatsep}{4pt plus 2pt minus 2pt}
\setlength{\intextsep}{6pt plus 2pt minus 2pt}
\setlength{\abovecaptionskip}{2pt}
\setlength{\belowcaptionskip}{0pt}
\setlength{\abovedisplayskip}{4pt plus 2pt minus 2pt}
\setlength{\belowdisplayskip}{4pt plus 2pt minus 2pt}
\setlength{\abovedisplayshortskip}{3pt plus 2pt minus 2pt}
\setlength{\belowdisplayshortskip}{3pt plus 2pt minus 2pt}
\setlength{\jot}{2pt}
\linespread{0.99}


\begin{acronym}
    \acro{MPC}[MPC]{Model Predictive Control}
    \acro{MRAV}[MRAV]{Multi-Rotor Aerial Vehicle}
    \acro{NLP}[NLP]{Nonlinear Programming}
    \acro{NMPC}[NMPC]{Nonlinear Model Predictive Control}
    \acro{RMSE}[RMSE]{Root Mean Square Error}
    \acro{UAV}[UAV]{Unmanned Aerial Vehicle}
    \acro{wrt}[w.r.t.]{with respect to}
\end{acronym}



\begin{abstract}

    This paper presents a sensitivity-based tube \ac{NMPC} framework for cooperative aerial chains under bounded parametric uncertainty. We consider a planar two-vehicle chain connected by rigid links, modeled with input-rate actuation to enforce slew-rate and magnitude limits on thrust and torque. Robustness to uncertainty in link mass, length, and inertia is achieved by propagating first-order parametric state sensitivities along the horizon and using them to compute online constraint-tightening margins. We robustify an inter-link separation constraint, implemented via a smooth cosine embedding, and thrust-magnitude bounds. The method is implemented in MATLAB and evaluated with boundary-hugging maneuvers and Monte-Carlo uncertainty sampling. Results show improved constraint margins under uncertainty with tracking performance comparable to nominal \ac{NMPC}.

\end{abstract}



\section{Introduction}
\label{sec:introduction}

\acfp{UAV} are widely used for inspection and monitoring thanks to rapid deployment and access to confined or otherwise hard-to-reach environments \cite{OlleroTRO2022, RuggieroRAL2018}. An increasing number of applications are inherently \emph{constraint-critical}, including close-range inspection \cite{OlleroTFR2025}, contact-aided sensing \cite{MalczykAURO2023}, and operation near extended assets \cite{RuggieroRAL2018}. In such tasks, safety and success depend on respecting geometric and actuation limits rather than solely tracking a nominal trajectory. Even short-lived constraint violations may be unacceptable.

Aerial physical interaction is commonly studied under the umbrella of \textit{aerial manipulation}. The term includes both aerial robots equipped with manipulators (e.g., arms or grippers) and cooperative multi-vehicle systems where interaction emerges from distributed thrust and internal-force regulation. Surveys \cite{OlleroTRO2022, RuggieroRAL2018} consistently identify constraint satisfaction under uncertainty as a key challenge for aerial manipulation, especially near obstacles and actuator limits and when the task involves orientation variables subject to periodicity.

\begin{figure}
    \centering
    \includegraphics[width=0.75\linewidth]{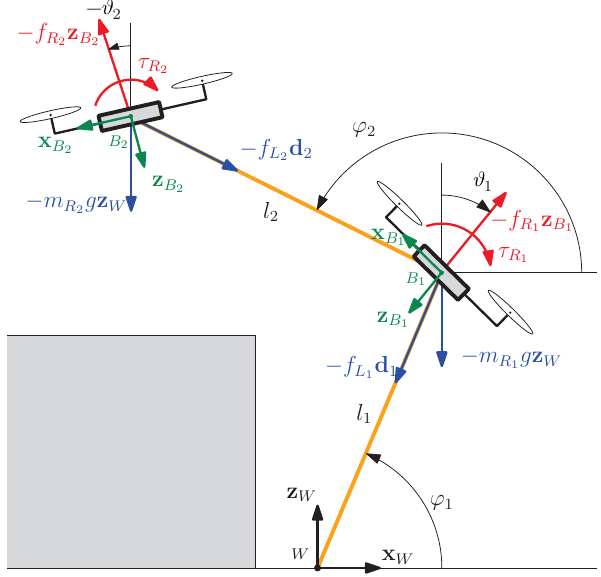}
    \vspace{-0.3em}
    \caption{Representation of the cooperative aerial structure and its main variables. The system is depicted in an example manipulation scenario, where the gray box represents a surface for, e.g., a pick-and-place task \cite{Tognon2015ECC}.}
    \label{fig:chainTwoUAVs}
\end{figure}

In this work, we focus on cooperative aerial structures \cite{OlleroTRO2022, KimTM2018}, where manipulation capability is achieved through actuation distributed across multiple aerial vehicles interconnected by rigid links. A minimal yet representative example is a chain of two underactuated aerial robots connected through a rigid bar \cite{Goodman2022JNS, Tognon2015ECC} (see Figure~\ref{fig:chainTwoUAVs}), whose control objectives typically include both motion regulation and internal-stress management. Larger-scale realizations, such as distributed-rotor aerial skeletons \cite{YangICRA2018}, further motivate control designs that can handle strong dynamic coupling and constraint activation in a systematic manner.

The same properties that make cooperative aerial structures attractive also make them difficult to control \cite{OlleroTRO2022, RuggieroRAL2018}. Their dynamics are nonlinear and strongly coupled: link geometry, internal forces, and thrust directions interact, and the overall system is underactuated. Moreover, unavoidable mismatch in structural parameters (e.g., effective mass, length, and inertia) can significantly alter the predicted evolution of constraint-relevant quantities. 
As a consequence, a nominal controller may satisfy constraints for the nominal model while violating them on the physical system \cite{GrunePannekNMPC2017, RawlingsMayneDiehlMPC2022}. This motivates robust \ac{MPC} \cite{Kohler2020IJRNC, LangsonAutomatica2004}: among its formulations, \textit{tube} \ac{MPC} \cite{LangsonAutomatica2004, LIMON200815333, Nyboe2025ICRA, JuACC2018} tightens constraints to keep the true system feasible for all admissible uncertainties.

For nonlinear systems, however, computing tubes exactly is generally intractable online, motivating practical approximations based on local linearization and sensitivity propagation \cite{Giordano2018ICRA}. Sensitivity-based tightening \cite{LeisterACC2025, Giordano2018ICRA, BelvedereTRO2025} provides a computationally efficient mechanism for estimating uncertainty-induced constraint variation, but its use for cooperative aerial chains with distributed actuation and nonlinear geometric constraints has not been explored, yet.

This paper develops a sensitivity-based tube \acf{NMPC} formulation for a planar cooperative aerial structure and studies robust constraint satisfaction under bounded parametric uncertainty in structural properties.
%
The main contributions are as follows. First, we adopt an \textit{input-rate} \ac{NMPC} formulation that enables direct enforcement of actuator slew-rate limits and ensures smooth actuation profiles. Second, we augment the prediction model with first-order parametric state sensitivities to propagate uncertainty effects online. Third, we use these sensitivities to compute constraint-tightening margins for nonlinear constraints, yielding a computationally tractable tube \ac{NMPC} formulation. Finally, we validate the approach through boundary-hugging maneuvers and Monte-Carlo simulations, demonstrating improved constraint robustness under parametric uncertainty while preserving tracking performance.

While the tightening mechanism applies to general nonlinear constraints, we focus on two representative classes. The first is an \textit{inter-link angular separation} constraint, which prevents physically undesirable near-aligned configurations and is naturally expressed using periodic functions of a relative angle; such constraints must be treated carefully in optimization-based controllers because non-smooth wrapping at branch cuts can compromise linearization-based bounds \cite{KALABIC2017293}. The second class comprises \textit{thrust-magnitude limits}, which are smooth but dynamically coupled constraints that often become active during aggressive maneuvers or under model mismatch, providing a complementary test case for the proposed tightening strategy.

The proposed approach is deterministic and targets bounded parametric uncertainty in structural properties of a cooperative aerial chain. Unlike min--max robust \ac{NMPC} formulations (see, e.g., \cite{RawlingsMayneDiehlMPC2022, GrunePannekNMPC2017}), it retains the computational structure of nominal \ac{NMPC} and introduces robustness through sensitivity propagation and constraint tightening, which can be implemented efficiently within standard nonlinear optimization frameworks.



\subsection{Related work}
\label{sec:relatedWork}

Cooperative aerial manipulation is an alternative to arm-equipped single-\ac{UAV} platforms when payload limits, actuation redundancy, or geometric scalability are critical \cite{OlleroTRO2022, RuggieroRAL2018}. A canonical benchmark is the two-\ac{UAV} rigid-link chain of \cite{Tognon2015ECC}, while distributed-rotor aerial skeletons such as LASDRA \cite{YangICRA2018} extend the same distributed-actuation principle to larger structures. Hence, we adopt the two-\ac{UAV} chain as a minimal archetype and focus on robustness to parametric uncertainty and constraint satisfaction.

\ac{NMPC} is a standard framework for nonlinear constrained systems \cite{GrunePannekNMPC2017, RawlingsMayneDiehlMPC2022} and has been applied to aerial robots with tight actuation limits and aggressive maneuvers, including input-rate formulations based on actuator dynamics \cite{BicegoJIRS2020}. It has also been demonstrated for online constrained end-effector tracking in aerial manipulation \cite{LunniICUAS2017}. In contrast to nominal formulations, we propagate parametric sensitivities online to enable constraint tightening.

Robust \ac{MPC} and tube-based formulations enforce constraint satisfaction under bounded uncertainty \cite{LangsonAutomatica2004, LIMON200815333}. For nonlinear systems, exact tube computation is generally intractable in real time, motivating approximations based on local linearization and sensitivity propagation \cite{Giordano2018ICRA}. Sensitivity-based tightening offers a practical trade-off between conservatism and tractability \cite{LeisterACC2025, Kohler2020IJRNC}. Accordingly, we study it for cooperative aerial chains with distributed actuation, internal-force coupling, and wrapped geometric constraints, and validate it under boundary-hugging maneuvers.



\section{System and Uncertainty Model}
\label{sec:modelingUncertainty}

\begin{table}[tb]
    \centering
    \caption{Notation summary.}
    \vspace{-0.25em}
    \label{tab:tubeNMPCSymbols}
    \begin{tabular}{ll}
    \toprule
    \textbf{Symbol} & \textbf{Description} \\
    \midrule
    $\mathbf{x}$, $\mathbf{x}_i$ & Extended state and its stage \\
    $\bar{\mathbf{x}}$ & Augmented state $[\mathbf{x}^\top,\mathrm{vec}(\bm{\Pi})^\top]^\top$ \\
    $\mathbf{u}$, $\mathbf{u}_i$ & Input rate (thrust/torque derivatives) and its stage\\
    $t_k$, $T_s$, $N$ & Sample time index, sampling period, horizon length \\
    $\mathbf{p}$, $\mathbf{p}_0$, $\mathcal{P}$ & Uncertain parameters, nominal value, uncertainty set \\
    $\bm{\Pi}$, $\bm{\Pi}_i$ & State sensitivities $\partial\mathbf{x}/\partial\mathbf{p}$ \\
    $\bm{\Pi}_y$ & Constraint sensitivity $\partial y/\partial\mathbf{p}$ \\
    $\alpha_y$, $\varepsilon_s$ & Tightening margin, regularization constant \\
    $y(\cdot)$, $y_{\max}$ & Constraint function and bound \\
    $\mathbf{W}_{\mathbf{p}}$ & Uncertainty weighting matrix \\
    $\mathbf{F}$, $\mathcal{F}_{\bm{\Pi}}$ & Discrete-time state and sensitivity maps \\
    $L(\cdot)$, $L_f(\cdot)$ & Stage/terminal costs \\
    $\mathbf{Q}$, $\mathbf{Q}_N$ & Output-tracking weights \\
    $\mathbf{h}(\cdot)$, $\mathbf{h}_N(\cdot)$ & Stage/terminal output maps \\
    $\mathbf{y}_i$, $\mathbf{y}_N$ & Predicted outputs \\
    $\mathbf{r}_i$, $\mathbf{r}_N$ & Stage/terminal references \\
    $\mathcal{X}$, $\mathcal{U}$ & Admissible state/input sets \\
    \bottomrule
    \end{tabular}
\end{table}

We consider a planar cooperative aerial structure composed of two underactuated aerial robots connected in series by two rigid links, i.e., a chain of two vehicles attached to a fixed point through links of constant lengths $\ell_1$ and $\ell_2$, as depicted in Figure~\ref{fig:chainTwoUAVs}. The setup follows the modeling line of \cite{Tognon2015ECC}: each link is connected to the corresponding vehicle center of mass through a passive joint, so that no rotational constraints are imposed at the attachment points. 

The system evolves on the vertical plane $(x_W,z_W)$ under four control inputs (two per robot). For each robot $j\in\{1,2\}$, the thrust magnitude $f_{R_j}\in\mathbb{R}_{>0}$ generates the force $-f_{R_j}\,\bm z_{B_j}$ along the body-fixed thrust direction $\bm z_{B_j}$, and the body torque magnitude $\tau_{R_j}\in\mathbb{R}$ generates the pitching moment $-\tau_{R_j}\,\bm y_{B_j}$ about the out-of-plane axis $\bm y_{B_j}$, where $\bm z_{B_j}$ and $\bm y_{B_j}$ denote body-frame unit vectors \cite{Tognon2015ECC}.

The configuration is described by the generalized coordinates $\mathbf{q} \triangleq [   \bm{\varphi}^\top, \bm{\vartheta}^\top]^\top = [\varphi_1, \varphi_2, \vartheta_1,  \vartheta_2]^\top \in \mathbb{R}^{4},$ where $\varphi_j \in \mathbb{R}$ denotes the elevation angle of the $j$-th link (measured \ac{wrt} the $x_W$ axis) and $\vartheta_j \in \mathbb{R}$ denotes the pitch angle of the $j$-th vehicle. Their time derivatives are denoted by $\dot{\varphi}_j$ and $\dot{\vartheta}_j$.

In addition, we denote by $f_{L_j}\in\mathbb{R}$ the internal force acting along the longitudinal direction of the $j$-th link (link stress), with $f_{L_j}>0$ corresponding to tension and $f_{L_j}<0$ to compression. In the considered model, $f_{L_j}$ is treated as an algebraic output obtained from the predicted state and actuation variables; it can therefore be constrained within the \ac{NMPC} framework when stress limits are imposed.

Unless otherwise stated, the modeling assumptions follow \cite{Tognon2015ECC}: (i) link masses and rotational inertias are negligible \ac{wrt} those of the vehicles (verified for lightweight carbon-fiber rods of mass $\ll m_{R_j}$), (ii) link lengths $\ell_j\in\mathbb{R}_{>0}$ are constant during operation (though their exact values may be uncertain, as modeled in Section~\ref{subsec:paramUncertainty}), and (iii) link deformations and elasticities are neglected. The full equations of motion are omitted for brevity and follow \cite{Tognon2015ECC}; in the remainder, we only require the resulting prediction model in control-oriented form, introduced after the input-rate augmentation. For convenience, the main symbols used in the tube \ac{NMPC} formulation are summarized in Table~\ref{tab:tubeNMPCSymbols}.



\subsection{Input-rate actuation and extended state}
\label{subsec:inputRate}

To explicitly enforce actuation slew-rate constraints within \ac{NMPC}, we adopt an input-extension strategy in which thrust and torque are treated as additional states and their time derivatives are used as optimization variables \cite{BicegoJIRS2020}. We define the extended state
\begin{equation} \label{eq:stateDef}
 \mathbf{x} \triangleq
\begin{bmatrix}
 \bm{\varphi}^\top, \bm{\vartheta}^\top, \dot{\bm{\varphi}}^\top, \dot{\bm{\vartheta}}^\top, \mathbf{f}_R^\top, \bm{\tau}_R^\top
\end{bmatrix}^\top \in \mathbb{R}^{12},
\end{equation}
where $\mathbf{f}_R \triangleq [f_{R_1}, f_{R_2}]^\top\in\mathbb{R}^{2}$ and $\bm{\tau}_R \triangleq [\tau_{R_1}, \tau_{R_2}]^\top\in\mathbb{R}^{2}$ collect the thrust and torque magnitudes of the two vehicles.

The \ac{NMPC} input is the actuation-rate vector $\mathbf{u} \triangleq  [\mathbf{u}_f^\top, \mathbf{u}_\tau^\top] = [\dot{\mathbf{f}}_R^\top, \dot{\bm{\tau}}_R^\top]^\top \in \mathbb{R}^4$, which allows imposing box constraints on $\dot{f}_{R_j}$ and $\dot{\tau}_{R_j}$ directly, while $f_{R_j}$ and $\tau_{R_j}$ remain available as state constraints (e.g., thrust saturation).

The resulting prediction model is $\dot{\mathbf{x}} = \mathbf{f}(\mathbf{x},\mathbf{u},\mathbf{p})$,
where $\mathbf{p}$ collects uncertain physical parameters. The vector field $\mathbf{f}(\cdot)$ is obtained by augmenting the nominal chain dynamics with the integrator relations $\dot{\mathbf{f}}_R = \mathbf{u}_f$ and $\dot{\bm{\tau}}_R = \mathbf{u}_\tau$. For the sensitivity-based analysis developed later, $\mathbf{f}(\cdot)$ is assumed continuously differentiable in the operating region of interest.



\subsection{Parametric uncertainty}
\label{subsec:paramUncertainty}

We consider bounded parametric uncertainty affecting link properties. Let $\mathbf{p} \triangleq [\bm{\delta}_m^\top, \bm{\delta}_\ell^\top, \bm{\delta}_J^\top]^\top \in \mathbb{R}^{6}$, where $\bm{\delta}_m = [\delta_{m_1},\delta_{m_2}]^\top$, $\bm{\delta}_\ell = [\delta_{\ell_1},\delta_{\ell_2}]^\top$, and $\bm{\delta}_J = [\delta_{J_1},\delta_{J_2}]^\top$ denote relative deviations from nominal values of representative parameters (mass $m_R$, length $\ell_R$, and inertia $J_R$). The corresponding physical parameters are modeled through multiplicative scalings $m_j=(1+\delta_{m_j})m_{R_j}$, $\ell_j=(1+\delta_{\ell_j})\ell_{R_j}$, and $J_j=(1+\delta_{J_j})J_{R_j}$, with $j \in \{1,2\}$.

The parameters $(m_{R_j},\ell_{R_j},J_{R_j})$ should be interpreted as effective quantities capturing structured model mismatch in the link-related dynamics; the assumptions of \cite{Tognon2015ECC} are recovered by setting $\delta_{m_j}=\delta_{J_j}=0$.

The admissible uncertainty set is the box
\begin{equation}\label{eq:uncSet}
\resizebox{0.910\hsize}{!}{$%
\mathcal{P} \triangleq 
\left\{\mathbf{p}: |\delta_{m_j}|\le \delta_{m_j,\max},\ |\delta_{\ell_j}|\le \delta_{\ell_j,\max},\ |\delta_{J_j}|\le \delta_{J_j,\max}\right\},
$}%
\end{equation}
which reflects relative error bounds typically available from identification and manufacturing tolerances.

The bounds $\delta_{m_j,\max}$, $\delta_{\ell_j,\max}$, and $\delta_{J_j,\max}$ are dimensionless relative deviations. For example, $\delta_{m_j,\max}=0.25$ represents a $\pm 25\%$ variation in the nominal link mass $m_{R_j} = \SI{0.457}{kg}$, which is consistent with typical uncertainty levels in aerial structures identified from experiments \cite{Tognon2015ECC}.



\section{Nominal NMPC Formulation}
\label{sec:NMPCFormulation}

At each sampling instant $t_k$ (with $t_k \triangleq kT_s$ and sampling period $T_s$), the controller solves a finite-horizon optimal control problem over a prediction horizon of $N$ steps. The predicted state sequence $\{\mathbf{x}_i\}_{i=0}^{N}$ corresponds to times $t_{k+i}=t_k+iT_s$. The prediction model is obtained by discretizing $\dot{\mathbf{x}}=\mathbf{f}(\mathbf{x},\mathbf{u},\mathbf{p})$ with a fixed-step integration scheme, yielding the discrete-time mapping $\mathbf{x}_{i+1}=\mathbf{F} (\mathbf{x}_i,\mathbf{u}_i,\mathbf{p}_0),$ with $i=0,\ldots,N-1$, where $\mathbf{p}_0$ denotes the nominal parameter vector ($\bm{\delta}_m = \bm{\delta}_\ell = \bm{\delta}_J = 0$) and $\mathbf{x}_0=\mathbf{x}(t_k)$ is the measured/estimated state.



\subsection{Finite-horizon optimal control problem}
\label{sec:finite-horizonOP}

The nominal \ac{NMPC} problem is formulated as
\begin{subequations}\label{eq:NMPC_nominal}
\begin{align}
\minimize_{\{\mathbf{x}_i,\mathbf{u}_i\}_{i=0}^{N-1}} \quad
& \sum_{i=0}^{N-1} L(\mathbf{x}_i,\mathbf{u}_i,\mathbf{r}_i) + L_f(\mathbf{x}_N,\mathbf{r}_N) \\[0.5mm]
\text{s.t.}\quad
& \mathbf{x}_{i+1}=\mathbf{F}(\mathbf{x}_i,\mathbf{u}_i,\mathbf{p}_0), \; i=0,\ldots,N-1,\\
& \mathbf{x}_i \in \mathcal{X},\ \mathbf{u}_i \in \mathcal{U}, \; i=0,\ldots,N-1,\\
& \mathbf{x}_0 = \mathbf{x}(t_k),
\end{align}
\end{subequations}
where $\mathcal{X}\subset\mathbb{R}^{12}$ and $\mathcal{U}\subset\mathbb{R}^{4}$ encode slew-rate limits, thrust/torque saturation, and physical bounds; reference sequences $\mathbf{r}_i$, $\mathbf{r}_N$ are assumed available at $t_k$.

We employ an output-tracking objective, where the tracked output is defined as $\mathbf{y}_i \triangleq \mathbf{h}(\mathbf{x}_i,\mathbf{u}_i),$ and $\mathbf{y}_N \triangleq \mathbf{h}_N(\mathbf{x}_N),$ where $\mathbf{h}(\cdot)$ collects the quantities to be tracked. 

In this work, we define the task specified in terms of desired link elevation angles and internal stresses. Accordingly, we choose
\begin{equation}\label{eq:stageOutput}
\resizebox{0.89\hsize}{!}{$%
\mathbf{h}(\mathbf{x},\mathbf{u}) \triangleq 
\begin{bmatrix}
    \bm{\varphi} \\
    \mathbf{f}_L(\mathbf{x}, \mathbf{u}) \\
    \dot{\bm{\varphi}} \\
    \ddot{\bm{\varphi}}(\mathbf{x}, \mathbf{u})
\end{bmatrix}\in\mathbb{R}^{8},
\quad
\mathbf{h}_N(\mathbf{x}) \triangleq 
\begin{bmatrix}
    \bm{\varphi} \\
    \mathbf{f}_L(\mathbf{x}, \mathbf{0})
\end{bmatrix}\in\mathbb{R}^{4}.
$}%
\end{equation}

Here, $\mathbf{f}_L$ denotes the link stress, computed as an algebraic function of the predicted state and actuation variables (as in \cite{Tognon2015ECC}), while $\ddot{\bm{\varphi}}$ is obtained from the model dynamics $\mathbf{f}(\cdot)$.

The reference sequences $\{\mathbf{r}_i\}_{i=0}^{N-1}$ and $\mathbf{r}_N$ comprise desired elevation angles, link stresses, and their respective time derivatives. The design of the reference generator is outside the scope of this paper, and the reference sequences are assumed available at each sampling instant.

Using diagonal weighting matrices $\mathbf{Q}\succeq 0$ and $\mathbf{Q}_N\succeq 0$, the stage and terminal costs are defined as $L(\mathbf{x}_i,\mathbf{u}_i,\mathbf{r}_i) \triangleq \tfrac{1}{2}\|\mathbf{h}(\mathbf{x}_i,\mathbf{u}_i)-\mathbf{r}_i\|_{\mathbf{Q}}^{2}$ and $L_f(\mathbf{x}_N,\mathbf{r}_N) \triangleq \tfrac{1}{2}\|\mathbf{h}_N(\mathbf{x}_N)-\mathbf{r}_N\|_{\mathbf{Q}_N}^{2}$.



\subsection{Nominal constraints of interest}
\label{sec:nominalConstraints}

Consistently with the problem statement in the Introduction, we focus on two representative classes of nominal constraints that are subsequently robustified through tube tightening.

\paragraph{Inter-link separation constraint} To prevent the second link from collapsing onto the first one and to avoid physically infeasible near-aligned configurations, we constrain the relative inter-link angle $\Delta\varphi \triangleq \varphi_2 - \varphi_1.$

Specifically, we require the magnitude of the wrapped relative angle to remain above a prescribed threshold $\Delta\varphi_{\min}\in(0,\pi)$, namely
\begin{equation} \label{eq:antiAlign}
|\mathrm{wrap}(\Delta\varphi)| \ge \Delta\varphi_{\min},
\end{equation}
where $\mathrm{wrap}(\cdot)$ maps angles to a principal interval, e.g., $[-\pi,\pi]$.

Direct angle wrapping introduces a non-smooth branch cut at $\Delta\varphi=\pm\pi$ that would compromise the Jacobian-based tightening in \eqref{eq:alphaDef}--\eqref{eq:tubeNMPC}.
To avoid this, the constraint~\eqref{eq:antiAlign} is implemented through the smooth periodic inequality $\cos(\Delta\varphi) \leq \cos(\Delta\varphi_{\min}),$
which provides an equivalent characterization of \eqref{eq:antiAlign} for $\Delta\varphi\in[-\pi,\pi]$ and $\Delta\varphi_{\min}\in(0,\pi)$, at the cost of a slight conservatism near the feasible-set boundary \cite{KALABIC2017293}. The resulting continuously differentiable constraint is enforced at all steps $i=0,\ldots,N$. 

\paragraph{Thrust-magnitude limits} Since the thrust magnitudes $\mathbf{f}_R=[f_{R_1}, f_{R_2}]^\top$ are included in the state vector, thrust saturation can be enforced directly through state constraints in $\mathcal{X}$. In particular, for each vehicle $j\in\{1,2\}$ and prediction step $i=0,\ldots,N$, we impose
\begin{equation}\label{eq:thrustBounds}
f_{R,\min} \le f_{R_{j},i} \le f_{R,\max}.
\end{equation}
Analogous bounds can be imposed on the torque states $\tau_{R_j}$ when actuator limits are considered. 

Sensitivity-based tightening is applied to both constraint classes above; torque magnitude limits are treated nominally.



\section{Sensitivity-Based Tube NMPC}
\label{sec:sensitivityTube}

Let $\mathbf{p}\in\mathbb{R}^{6}$ denote the vector of uncertain physical parameters introduced in Section~\ref{subsec:paramUncertainty}, namely the relative deviations in representative mass, length, and inertia. To quantify how such parametric uncertainty affects the predicted system evolution, we introduce the \emph{parametric state sensitivity matrix} $\bm{\Pi}(t) \triangleq \partial\mathbf{x}(t)/\partial\mathbf{p} \in\mathbb{R}^{12\times 6}$ \cite{Giordano2018ICRA}, whose columns represent the first-order variation of the state \ac{wrt} each uncertain parameter.

We assume the standard local smoothness condition: $\mathbf{f}(\cdot)$ is continuously differentiable in a neighborhood of the nominal trajectory generated using $\mathbf{p}_0$.\footnote{For the considered rigid-link aerial chain, this assumption requires predicted trajectories to remain away from kinematic singularities (e.g., near-aligned link configurations, where the kinematic relationships become ill-conditioned) and to involve only smooth actuator dynamics (including the cosine-based wrap constraint). Since the construction is local, violations may render the first-order sensitivity-based tube approximation unreliable.} Differentiating the dynamics $\dot{\mathbf{x}}=\mathbf{f}(\mathbf{x},\mathbf{u},\mathbf{p})$ \ac{wrt} $\mathbf{p}$ yields the standard first-order variational equation
\begin{equation}\label{eq:PiDyn}
\dot{\bm{\Pi}}(t) = \mathbf{f}_\mathbf{x}(\mathbf{x}(t),\mathbf{u}(t),\mathbf{p}_0)\,\bm{\Pi}(t) + \mathbf{f}_\mathbf{p}(\mathbf{x}(t),\mathbf{u}(t),\mathbf{p}_0),
\end{equation}
where $\mathbf{f}_\mathbf{x} \triangleq \partial \mathbf{f} / \partial \mathbf{x},$ and $\mathbf{f}_\mathbf{p} \triangleq \partial  \mathbf{f}/\partial \mathbf{p}$.

If the initial state estimate is assumed independent of $\mathbf{p}$, the sensitivities can be initialized as $\bm{\Pi}(t_0)=\mathbf{0}$ at the start of the experiment.

In a receding-horizon implementation, $\bm{\Pi}(t_k)$ can then be treated as an internal state: it is propagated between sampling instants using~\eqref{eq:PiDyn} (or its discrete-time counterpart, consistently with the numerical integration used in the \ac{NMPC} transcription) and used to initialize each \ac{NMPC} solve, i.e., $\bm{\Pi}_0=\bm{\Pi}(t_k)$ as in~\eqref{eq:tube_init}.



\subsection{Augmented prediction model}
\label{subsec:augmentedModel}

To propagate state sensitivities along the \ac{NMPC} prediction horizon together with the nominal system dynamics, we augment the state vector with the vectorized sensitivity matrix. Specifically, we define the augmented state $\bar{\mathbf{x}} \triangleq [\mathbf{x}^\top\ \mathrm{vec}(\bm{\Pi})^\top]^\top \in \mathbb{R}^{12+72},$ where $\mathrm{vec}(\cdot)$ stacks the columns of $\bm{\Pi}$ into a single vector.

The corresponding augmented continuous-time dynamics are given by
\begin{equation}\label{eq:xAugDyn}
\dot{\bar{\mathbf{x}}} =
\begin{bmatrix}
\mathbf{f}(\mathbf{x},\mathbf{u},\mathbf{p}_0)\\
\mathrm{vec}\!\big(\mathbf{f}_\mathbf{x}(\mathbf{x},\mathbf{u},\mathbf{p}_0)\,\bm{\Pi} + \mathbf{f}_\mathbf{p}(\mathbf{x},\mathbf{u},\mathbf{p}_0)\big)
\end{bmatrix}.
\end{equation}

This formulation allows the \ac{NMPC} optimizer to simultaneously predict the nominal trajectory and the associated parametric sensitivities at every stage of the horizon.

In the implementation, the Jacobians $\mathbf{f}_{\mathbf{x}}$ and $\mathbf{f}_{\mathbf{p}}$
are obtained by automatic differentiation and discretized consistently with the same integration scheme used for the nominal prediction model.



\subsection{Tightening margins and tube NMPC problem}
\label{subsec:tighteningMargins}

For a generic scalar constraint $y(\mathbf{x},\mathbf{p})\le y_{\max}$. Along a nominal prediction computed with $\mathbf{p}_0$, the first-order sensitivity of the constraint function $y$ \ac{wrt} the uncertain parameters $\mathbf{p}$ is given by
\begin{equation}\label{eq:ySensDef}
\bm{\Pi}_y \triangleq \frac{\partial y}{\partial \mathbf{p}} = \mathbf{J}_{y\mathbf{x}}\,\bm{\Pi} + \mathbf{J}_{y\mathbf{p}},
\end{equation}
where $\mathbf{J}_{y\mathbf{x}}\triangleq \partial y/\partial \mathbf{x}$ and $\mathbf{J}_{y\mathbf{p}}\triangleq \partial y/\partial \mathbf{p}$. For purely state-dependent constraints $y(\mathbf{x})$, one has $\mathbf{J}_{y\mathbf{p}}=\mathbf{0}$.

Given a bounded uncertainty set $\mathcal{P}$, we approximate the worst-case first-order variation of the constraint value by the scalar margin
\begin{equation}\label{eq:alphaDef}
\alpha_y \triangleq \sqrt{\bm{\Pi}_y\,\mathbf{W}_\mathbf{p}\,\bm{\Pi}_y^\top}+\varepsilon_s,
\end{equation}
where $\mathbf{W}_{\mathbf{p}}\succeq 0$ encodes the size and geometry of the uncertainty set,
and $\varepsilon_s>0$ is a small regularization constant added for numerical robustness.

For independent bounded parameters with box-type uncertainty set~\eqref{eq:uncSet}, a convenient choice is $\mathbf{W}_{\mathbf{p}}= \mathrm{diag}(\delta_{m_1,\max}^2, \delta_{m_2,\max}^2, \delta_{\ell_1,\max}^2, \delta_{\ell_2,\max}^2, \delta_{J_1,\max}^2, \delta_{J_2,\max}^2)$, which outer-approximates $\mathcal{P}$ with an ellipsoid.

Within the sensitivity-based tube \ac{NMPC} formulation, each nominal constraint
$y(\mathbf{x}_i,\mathbf{p}_0)\le y_{\max}$ is replaced at prediction stage $i$ by the tightened form
\begin{equation}\label{eq:tightenedConstraint}
 y(\mathbf{x}_i,\mathbf{p}_0) + \alpha_y(\mathbf{x}_i,\bm{\Pi}_i) \le y_{\max}, \quad i=0,\ldots,N.
\end{equation}

\textit{Interpretation:} The margin~\eqref{eq:alphaDef} is accurate in mildly nonlinear regimes; empirical validation for large deviations is provided via Monte-Carlo in Section~\ref{sec:simulationResults}.

At each sampling instant $t_k$, the tube \ac{NMPC} solves

\vspace*{-1.0em}
    \begin{subequations}\label{eq:tubeNMPC}
    \small
    \begin{align}
    \minimize_{\{\mathbf{x}_i,\mathbf{u}_i,\bm{\Pi}_i\}_{i=0}^{N-1}} \quad
    & \sum_{i=0}^{N-1} L(\mathbf{x}_i,\mathbf{u}_i,\mathbf{r}_i) + L_f(\mathbf{x}_N,\mathbf{r}_N) \label{eq:tube_cost}\\
    &\hspace{-5em}\text{s.t.}\quad \mathbf{x}_{i+1} = \mathbf{F}(\mathbf{x}_i,\mathbf{u}_i,\mathbf{p}_0), \ i=0,\ldots,N-1, \label{eq:tube_dyn}\\
    &\hspace{-3em} \bm{\Pi}_{i+1} = \mathcal{F}_{\bm{\Pi}}(\mathbf{x}_i,\mathbf{u}_i,\bm{\Pi}_i,\mathbf{p}_0), \ i=0,\ldots,N-1, \label{eq:tube_sens}\\
    &\hspace{-3em} \mathbf{x}_i \in \mathcal{X},\ \mathbf{u}_i \in \mathcal{U}, \ i=0,\ldots,N-1, \label{eq:tube_sets}\\
    &\hspace{-3em} y(\mathbf{x}_i,\mathbf{p}_0) + \alpha_y(\mathbf{x}_i,\bm{\Pi}_i) \le y_{\max}, \ i=0,\ldots,N, \label{eq:tube_tight}\\
    &\hspace{-3em} \mathbf{x}_0 = \mathbf{x}(t_k), \ \bm{\Pi}_0 = \bm{\Pi}(t_k). \label{eq:tube_init}
    \end{align}
    \normalsize
\end{subequations}

where $\mathcal{F}_{\bm{\Pi}}$ is the discrete-time sensitivity update map obtained by discretizing~\eqref{eq:PiDyn} consistently with $\mathbf{F}$.



\section{Simulation Results}
\label{sec:simulationResults}

Nominal and tube \ac{NMPC} are implemented with identical horizons, discretization, weights, and solver settings to ensure a fair comparison. The prediction horizon is set to $N=30$ with sampling time $T_s=\SI{0.01}{\second}$, corresponding to a \SI{0.3}{\second} prediction window. The dynamics are discretized using a fixed-step fourth-order Runge--Kutta integrator within a multiple-shooting formulation. The resulting nonlinear programs are solved in MATLAB using \textsc{MATMPC}\footnote{\url{https://github.com/chenyutao36/MATMPC}} in real-time iteration mode with \textsc{qpOASES}\footnote{\url{https://github.com/coin-or/qpOASES}}.

The reference trajectory is generated at \SI{200}{\hertz} and provided to the controller as a time-varying output reference. The \ac{NMPC} controller runs at \SI{100}{\hertz}. The closed-loop plant dynamics are integrated at \SI{200}{\hertz}.

\begin{figure}
    \centering
    \adjincludegraphics[width=0.75\columnwidth, trim={{.07\width} {.05\height} {.08\width} {.06\height}}, clip]{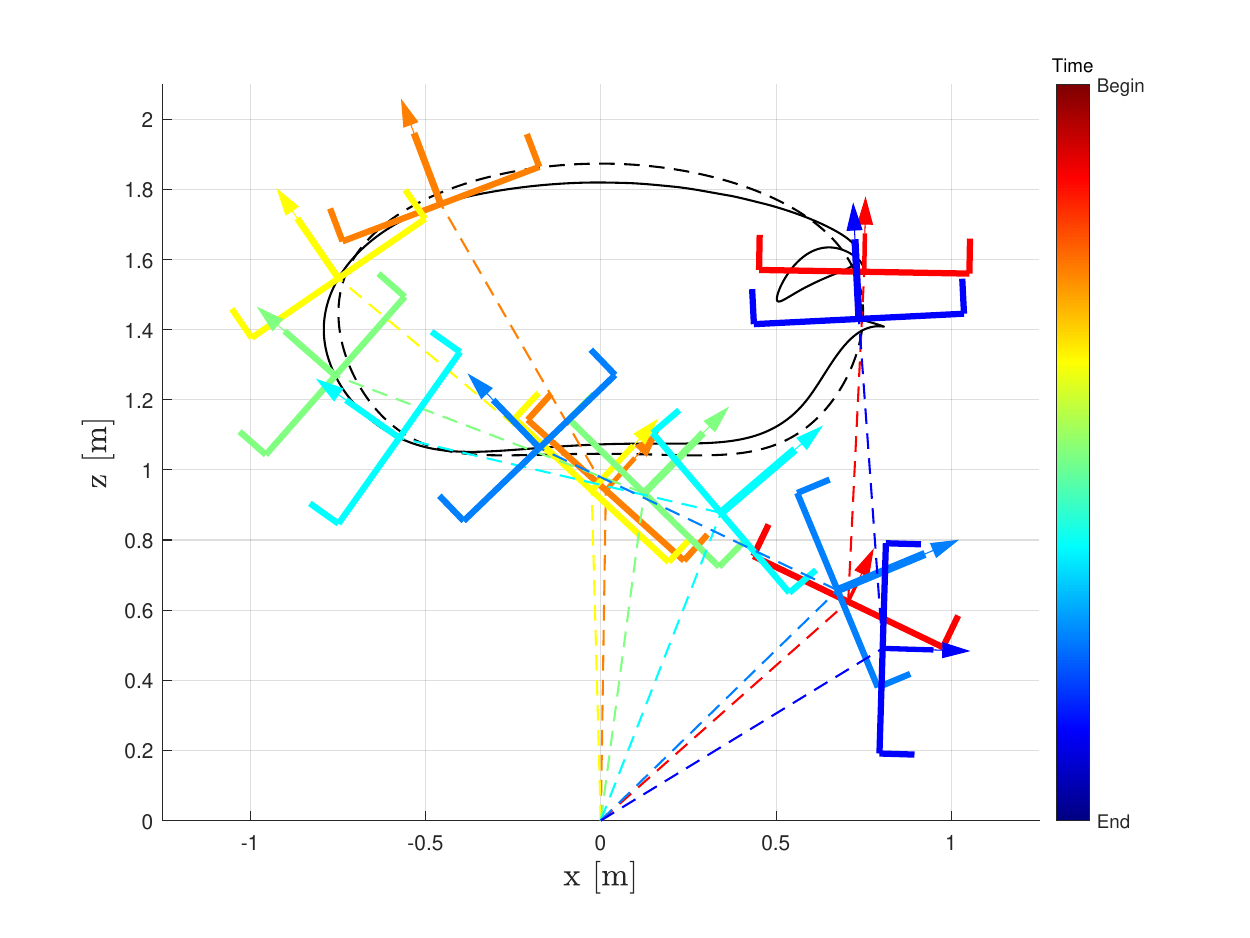}
    \vspace{-0.25em}
    \caption{Boundary-hugging maneuver under parametric uncertainty of the nominal \ac{NMPC}. Snapshots of the planar two-link aerial chain from $t=\SI{0}{s}$ to $t=\SI{12}{s}$. The dashed curve is the reference end-effector path, and the solid curve is the executed end-effector path.}
    \label{fig:2DNominalNMPC}
\end{figure}

\begin{figure}
    \centering
    \adjincludegraphics[width=0.75\columnwidth, trim={{.07\width} {.05\height} {.08\width} {.06\height}}, clip]{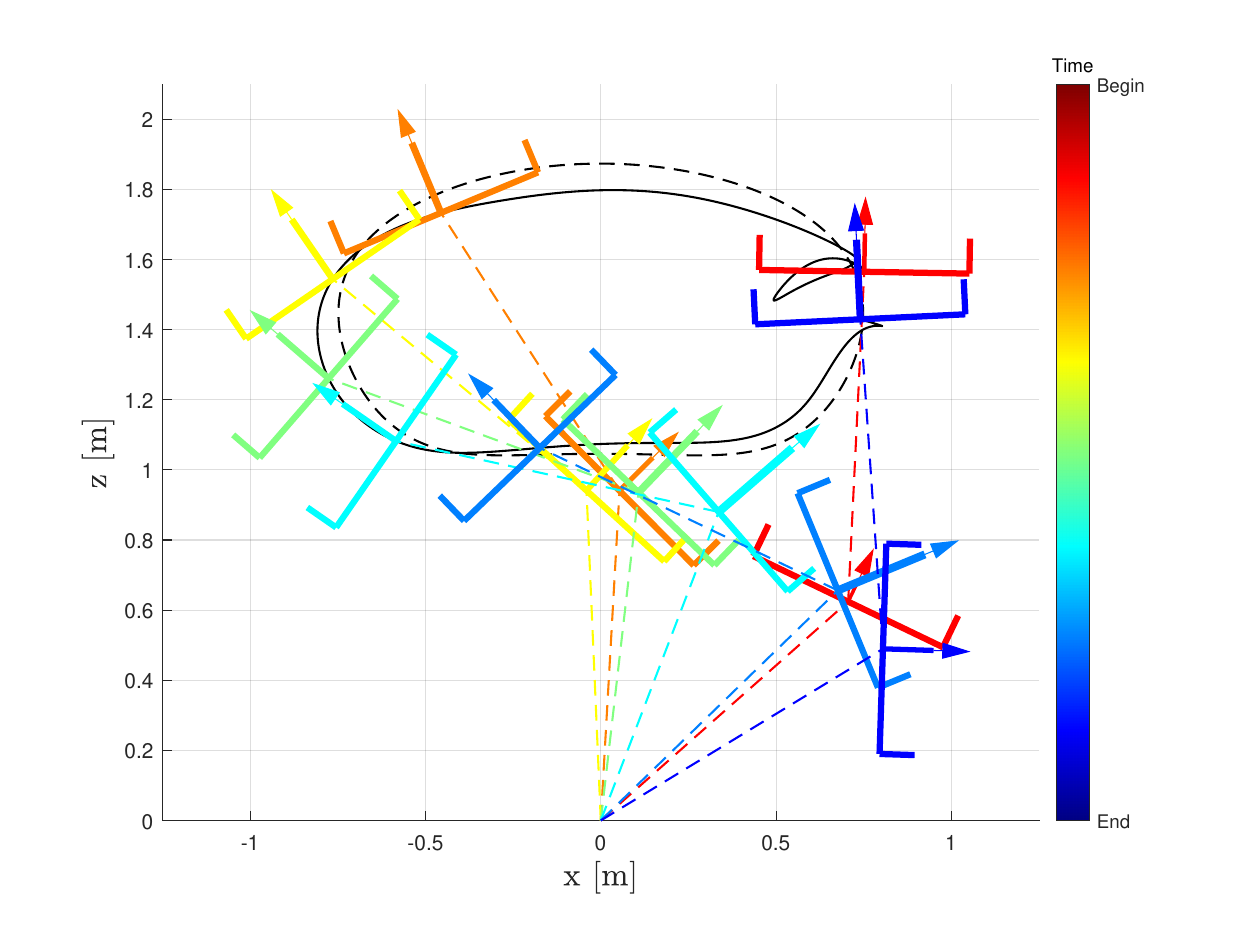}
    \vspace{-0.25em}
    \caption{Boundary-hugging maneuver under parametric uncertainty of the tube \ac{NMPC}.}
    \label{fig:2DTubeNMPC}
\end{figure}

To stress the controller near constraint boundaries, we use a \textit{boundary-hugging} reference trajectory defined by the ellipse $x_d(t) = x_c + a_x \cos\!\big(\nu(t)\big)$ and $z_d(t) = z_c + a_z \sin\!\big(\nu(t)\big),$ with semi-axes $a_x,$ $a_z$ and center $(x_c,z_c),$ designed so that small parameter deviations readily trigger constraint activation. The phase $\nu(t)$ is generated using a \textit{quintic time-scaling law}, i.e., a fifth-degree polynomial scheduling of the phase from $\nu(0)=0$ to $\nu(T)=2\pi$ that enforces zero initial/final velocity and acceleration. The corresponding elevation-angle references $\bm{\varphi}^d$ are obtained via inverse kinematics \cite{Tognon2015ECC}. Parameters are chosen so that $\rho^d(t) = \sqrt{(x^d)^2 + (z^d)^2}$ stays near the reachable boundary $\rho^d\in[0,2\ell]$ with margin $\varepsilon_r$ (see Table~\ref{tab:parameters}).
The desired link stress is set to a constant tension $f_L^d=\SI{10}{N}$ for both links. To introduce a realistic non-nominal initialization, the initial elevations are perturbed by $\mathbf{e}_{\varphi,0}=[-8^\circ,\;4^\circ]^\top$ \ac{wrt} the initial reference. Figures~\ref{fig:2DNominalNMPC} and~\ref{fig:2DTubeNMPC} illustrate the boundary-hugging maneuver by showing snapshots of the aerial chain at multiple instants along the motion, from the initial to the final configuration, together with the corresponding end-effector path, for nominal and tube \ac{NMPC}. 

Unless otherwise stated, tightening applies to the inter-link separation and thrust-magnitude constraints; torque limits are enforced nominally.

\begin{table}[tb]
    \centering
    \caption{System, uncertainty, and \ac{NMPC} parameters.}
    \vspace{-0.25em}
    \label{tab:parameters}
    \renewcommand{\arraystretch}{1.1}
    \begin{tabular}{lc|lc}
        \hline
        \multicolumn{2}{c|}{\textbf{Physical parameters}} &
        \multicolumn{2}{c}{\textbf{Reference Trajectory}} \\
        \hline
        $m_R$              & $0.457~\mathrm{kg}$      &
        $\Delta\varphi_{\min}$    & $30^\circ$ \\
        $J_R$              & $0.123~\mathrm{kg\,m^2}$ &
        $dt$               & $0.005~\mathrm{s}$ \\
        $\ell$             & $0.942~\mathrm{m}$      &
        $T$                & $12~\mathrm{s}$ \\
        $g$                & $9.81~\mathrm{m/s^2}$   &
        $f_L^{d}$          & $10~\mathrm{N}$ \\
        $n$                & $2~[-]$                     & 
        $\mathbf{e}_{\varphi,0}$    & $[-8^\circ,\;4^\circ]$ \\
        $f_{R,\min},f_{R,\max}$       & $\{3,20\}~\mathrm{N}$       & 
        $x_c, z_c$         & $\{0,1.05\}~\mathrm{m}$ \\
        $\tau_{R,\min},\tau_{R,\max}$    & $\{-5, 5\}~\mathrm{N\,m}$    & 
        $a_x, a_z$         & $\{0.55,0.35\}~\mathrm{m}$ \\
        $\dot{f}_{R,\min},\dot{f}_{R,\max}$ & $\{-200,200\}~\mathrm{N/s}$ &
        $\varepsilon_r$    & $0.02~\mathrm{m}$\\
        $\dot{\tau}_{R,\min},\dot{\tau}_{R,\max}$ & $\{-100,100\}~\mathrm{Nm/s}$ &
        --                 & -- \\
        \hline
        \multicolumn{2}{c|}{\textbf{\ac{NMPC} parameters}} & \multicolumn{2}{c}{\textbf{Uncertainty bounds}} \\
        \hline
        $T_s$              & $0.01~\mathrm{s}$            & $\delta_{m_j,\max}$    & $0.25~[-]$ \\
        $N$                & $30~[-]$                         & $\delta_{\ell_j,\max}$ & $0.24~[-]$ \\
        $\mathbf{Q}$       & $\mathrm{diag}(5,1,0.1,0.1)$ & $\delta_{J_j,\max}$    & $0.25~[-] $ \\
        $\mathbf{Q}_N$     & $\mathrm{diag}(50,10)$       & $\varepsilon_s$        & $10^{-12}~[-]$ \\
        $\varepsilon_{\mathrm{tol}}$ & $10^{-3}~[-]$ & $N_\mathrm{sim}$ & $500~[-]$ \\
        \hline
    \end{tabular}
\end{table}

\textbf{Single run.} We compare tracking performance and constraint satisfaction of nominal \ac{NMPC} and sensitivity-based tube \ac{NMPC} under parametric uncertainty.

\begin{figure}[tb]
    \centering
    \input{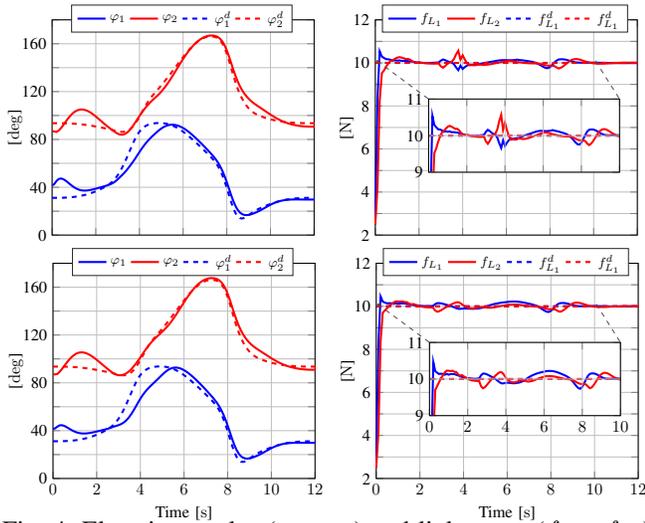}
    \vspace{-0.5em}
    \caption{Elevation angles ($\varphi_1,\ \varphi_2$) and link-stress ($f_{L_1},\ f_{L_2}$) for nominal \ac{NMPC} (top) and tube \ac{NMPC} (bottom). Dashed lines denote the references, while solid lines denote the closed-loop responses.}
    \label{fig:elevationLinkStress}
\end{figure}

\begin{figure}[tb]
    \centering
\hspace{-1.175cm}
\begin{subfigure}{0.425\columnwidth}
    \hspace*{0.095cm}
    \centering
    \scalebox{0.62}{
    \begin{tikzpicture}
    \begin{axis}[%
    width=2.2119in,%
    height=1.8183in,%
    at={(0.758in,0.481in)},%
    scale only axis,%
    xmin=0,%
    xmax=12.0,%
    ymax=25,%
    ymin=-25,%
    xmajorgrids,%
    ymajorgrids,%
    ylabel style={yshift=-0.455cm, xshift=0cm}, 
    extra x tick labels={ , ,  , },%
    ylabel={[\si{\deg}]},%
    ytick={-20,-10,0,10,20},%
    yticklabels={-20,-10,0,10,20},%
    xtick={0,2,...,12},%
    xticklabels={ , , },%
    axis background/.style={fill=white},%
    legend style={at={(0.495,0.17)},anchor=north,legend cell align=left,draw=none,legend 
    columns=-1,align=left,draw=white!15!black}
    ]
    \addplot [color=blue, solid, line width=1.25pt] 
        file{matlabPlots/elevation-linkStress/nominal/e_phi1.txt};%
    \addplot [color=red, solid, line width=1.25pt] 
        file{matlabPlots/elevation-linkStress/nominal/e_phi2.txt};%
    \legend{\footnotesize{$e_{\varphi_1}$}, \footnotesize{$e_{\varphi_2}$}};%
    \end{axis}
    \end{tikzpicture}
    }
\end{subfigure}
\hspace*{0.30cm}
\begin{subfigure}{0.425\columnwidth}
    \hspace*{0.095cm}
    \centering
    \scalebox{0.62}{
    \begin{tikzpicture}
    \begin{groupplot}[%
    group style={
            group size=2 by 1,
            horizontal sep=-11.5em, 
            vertical sep=0.0em, 
    },
    width=2.2119in,%
    height=1.8183in,%
    at={(0.758in,0.481in)},%
    scale only axis,%
    xmin=0,%
    xmax=12.0,%
    ymax=8.0,%
    ymin=-1,%
    xmajorgrids,%
    ymajorgrids,%
    ylabel style={yshift=-0.155cm, xshift=0cm}, 
    ylabel={[\si{\newton}]},%
    ytick={0,1,2,3,4,5,6,7,8,9,10,11,12},%
    yticklabels={0,\empty,2,\empty,4,\empty,6,\empty,8,\empty,10,\empty,12},%
    xtick={0,2,...,12},%
    xticklabels={ , , },%
    axis background/.style={fill=white},%
    legend style={at={(0.495,0.97)},anchor=north,legend cell align=left,draw=none,legend 
    columns=-1,align=left,draw=white!15!black}
    ]
    \nextgroupplot 
    \addplot [color=blue, solid, line width=1.25pt] 
        file{matlabPlots/elevation-linkStress/nominal/e_fL1.txt};%
    \addplot [color=red, solid, line width=1.25pt] 
        file{matlabPlots/elevation-linkStress/nominal/e_fL2.txt};%
    \coordinate (c1) at (axis cs:0.069,0);
    \coordinate (c2) at (axis cs:0.069,0.1);
    \coordinate (c3) at (axis cs:0.08,0);
    \coordinate (c4) at (axis cs:10.08,0.1);
    \draw[dashed, black!70] (c1) rectangle (axis cs:0.18,0.1);
    \legend{\footnotesize{$e_{f_{L_1}}$}, \footnotesize{$e_{f_{L_2}}$}};%
    \nextgroupplot[
        name = ax2,%
        width=1.6119in,%
        height=0.6183in,%
        scale only axis,%
        at={(0.45in,0.831in)},%
        anchor=south west,
        xmin=0,
        xmax=10,
        ymin=-1,
        ymax=1,
        xmajorgrids,%
        ymajorgrids,%
        ylabel style={yshift=-0.5cm, font=\footnotesize},
        ylabel={},
        xlabel={},
        xtick={0,2,4,6,8,10},
        yticklabels={-1,0,1},
        ytick={-1,0,1},
        area style,%
        title = { },%
        axis background/.style={fill=white},
    ] 
    \addplot [color=blue, solid, line width=1.25pt] 
        file{matlabPlots/elevation-linkStress/nominal/e_fL1.txt};%
    \addplot [color=red, solid, line width=1.25pt] 
        file{matlabPlots/elevation-linkStress/nominal/e_fL2.txt};%
    \end{groupplot}
    \draw [dashed, black!70] (c1) -- (ax2.south west);
    \draw [dashed, black!70] (c4) -- (ax2.south east);
    \end{tikzpicture}
    }
\end{subfigure}
\\
\vspace{0.025cm}
\hspace{-0.905cm}
\begin{subfigure}{0.425\columnwidth}
    \hspace*{0.095cm}
    \centering
    \scalebox{0.62}{
    \begin{tikzpicture}
    \begin{axis}[%
    width=2.2119in,%
    height=1.8183in,%
    at={(0.758in,0.481in)},%
    scale only axis,%
    xmin=0,%
    xmax=12.0,%
    ymax=25,%
    ymin=-25,%
    xmajorgrids,%
    ymajorgrids,%
    ylabel style={yshift=-0.455cm, xshift=0cm}, 
    xlabel={Time [\si{\second}]},%
    extra x tick labels={ , ,  , },%
    ylabel={[\si{\deg}]},%
    ytick={-20,-10,0,10,20},%
    yticklabels={-20,-10,0,10,20},%
    xtick={0,2,...,12},%
    axis background/.style={fill=white},%
    legend style={at={(0.495,0.17)},anchor=north,legend cell align=left,draw=none,legend 
    columns=-1,align=left,draw=white!15!black}
    ]
    \addplot [color=blue, solid, line width=1.25pt] 
        file{matlabPlots/elevation-linkStress/tube/e_phi1.txt};%
    \addplot [color=red, solid, line width=1.25pt] 
        file{matlabPlots/elevation-linkStress/tube/e_phi2.txt};%
    \legend{\footnotesize{$e_{\varphi_1}$}, \footnotesize{$e_{\varphi_2}$}};%
    \end{axis}
    \end{tikzpicture}
    }
\end{subfigure}
\hspace*{0.30cm}
\begin{subfigure}{0.425\columnwidth}
    \hspace*{0.095cm}
    \centering
    \scalebox{0.62}{
    \begin{tikzpicture}
    \begin{groupplot}[%
    group style={
            group size=2 by 1,
            horizontal sep=-11.5em, 
            vertical sep=0.0em, 
    },
    width=2.2119in,%
    height=1.8183in,%
    at={(0.758in,0.481in)},%
    scale only axis,%
    xmin=0,%
    xmax=12.0,%
    ymax=8.0,%
    ymin=-1,%
    xmajorgrids,%
    ymajorgrids,%
    ylabel style={yshift=-0.155cm, xshift=0cm}, 
    xlabel={Time [\si{\second}]},%
    ylabel={[\si{\newton}]},%
    ytick={0,1,2,3,4,5,6,7,8,9,10,11,12},%
    yticklabels={0,\empty,2,\empty,4,\empty,6,\empty,8,\empty,10,\empty,12},%
    xtick={0,2,...,12},%
    axis background/.style={fill=white},%
    legend style={at={(0.495,0.97)},anchor=north,legend cell align=left,draw=none,legend 
    columns=-1,align=left,draw=white!15!black}
    ]
    \nextgroupplot 
    \addplot [color=blue, solid, line width=1.25pt] 
        file{matlabPlots/elevation-linkStress/tube/e_fL1.txt};%
    \addplot [color=red, solid, line width=1.25pt] 
        file{matlabPlots/elevation-linkStress/tube/e_fL2.txt};%
    \coordinate (c1) at (axis cs:0.069,0);
    \coordinate (c2) at (axis cs:0.069,0.1);
    \coordinate (c3) at (axis cs:0.08,0);
    \coordinate (c4) at (axis cs:10.08,0.1);
    \draw[dashed, black!70] (c1) rectangle (axis cs:0.18,0.1);
    \legend{\footnotesize{$e_{f_{L_1}}$}, \footnotesize{$e_{f_{L_2}}$}};%
    \nextgroupplot[
        name = ax2,%
        width=1.6119in,%
        height=0.6183in,%
        scale only axis,%
        at={(0.45in,0.831in)},%
        anchor=south west,
        xmin=0,
        xmax=10,
        ymin=-1,
        ymax=1,
        xmajorgrids,%
        ymajorgrids,%
        ylabel style={yshift=-0.5cm, font=\footnotesize},
        ylabel={},
        xlabel={},
        xtick={0,2,4,6,8,10},
        yticklabels={-1,0,1},
        ytick={-1,0,1},
        area style,%
        title = { },%
        axis background/.style={fill=white},
    ] 
    \addplot [color=blue, solid, line width=1.25pt] 
        file{matlabPlots/elevation-linkStress/tube/e_fL1.txt};%
    \addplot [color=red, solid, line width=1.25pt] 
        file{matlabPlots/elevation-linkStress/tube/e_fL2.txt};%
    \end{groupplot}
    \draw [dashed, black!70] (c1) -- (ax2.south west);
    \draw [dashed, black!70] (c4) -- (ax2.south east);
    \end{tikzpicture}
    }
\end{subfigure}
    \vspace{-0.5em}
    \caption{Tracking errors for the trajectories in Figure~\ref{fig:elevationLinkStress}: nominal \ac{NMPC} (top) and tube \ac{NMPC} (bottom). Errors are defined as $e_{\varphi_j}\triangleq \varphi_j-\varphi_j^d$ and $e_{f_{L_j}}\triangleq f_{L_j}-f_{L_j}^d$, $j\in\{1,2\}$.}
    \label{fig:trackingErrors}
\end{figure}
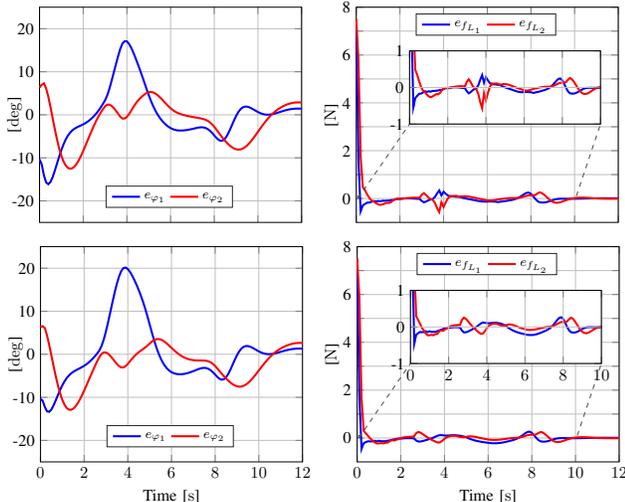

Figure~\ref{fig:elevationLinkStress} reports the elevation angles and link-stress outputs together with their references, while Figure~\ref{fig:trackingErrors} shows the corresponding tracking errors. 
In the representative run shown, the nominal and tube \ac{NMPC} closed-loop trajectories are visually very similar and yield comparable tracking accuracy. This outcome is expected when the particular uncertainty realization is mild and the tightened constraints are not persistently active: in that regime, the tube tightening mainly reshapes the feasible set near constraint-critical configurations, while leaving the unconstrained tracking optimum essentially unchanged.

The transient tracking error visible near $t=\SI{4}{s}$ in Figure~\ref{fig:trackingErrors} is primarily attributable to the initial elevation perturbation $\mathbf{e}_{\varphi,0}$ combined with the conservative tightening during the first constraint-active phase; it could be reduced by tuning the initial sensitivity $\bm{\Pi}_0$ or applying a shorter initial tightening horizon.

The main benefit of tube \ac{NMPC} should therefore not be inferred solely from output tracking plots, but rather from constraint-margins under uncertainty. The tightening increases safety margins in regions of high parametric sensitivity, reducing the risk of constraint activation under uncertainty.

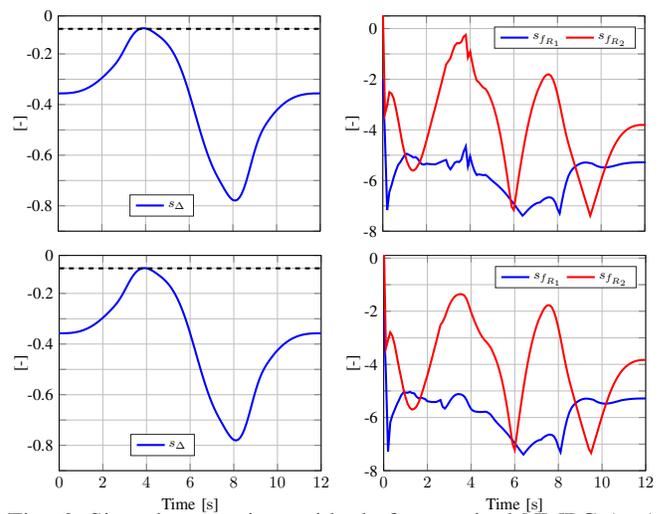
\begin{figure}[tb]
    \centering
\hspace{-1.175cm}
\begin{subfigure}{0.425\columnwidth}
    \hspace*{0.095cm}
    \centering
    \scalebox{0.62}{
    \begin{tikzpicture}
    \begin{axis}[%
    width=2.2119in,%
    height=1.8183in,%
    at={(0.758in,0.481in)},%
    scale only axis,%
    xmin=0,%
    xmax=12.0,%
    ymax=0.1,%
    ymin=-1.6,%
    xmajorgrids,%
    ymajorgrids,%
    ylabel style={yshift=-0.255cm, xshift=0cm}, 
    extra x tick labels={ , ,  , },%
    ylabel={[-]},%
    ytick={0.1,0,-0.2,...,-1.6},%
    yticklabels={0,\empty,-0.2,\empty,-0.4,\empty,-0.6,\empty,-0.8,\empty,1.0,\empty,-1.2,\empty,-1.4,\empty,-1.6},%
    xtick={0,2,...,12},%
    xticklabels={ , , },%
    axis background/.style={fill=white},%
    legend style={at={(0.395,0.17)},anchor=north,legend cell align=left,draw=none,legend 
    columns=-1,align=left,draw=white!15!black}
    ]
    \addplot [color=blue, solid, line width=1.25pt] 
        file{matlabPlots/elevation-linkStress/nominal/s_delta.txt};%
    \draw [line width=1.25pt, dashed] (rel axis cs: 0, 0.94) -- (rel axis cs: 12, 0.94);
    \legend{\footnotesize{$s_{\Delta}$}};%
    \end{axis}
    \end{tikzpicture}
    }
\end{subfigure}
\hspace*{0.50cm}
\begin{subfigure}{0.425\columnwidth}
    \hspace*{0.095cm}
    \centering
    \scalebox{0.62}{
    \begin{tikzpicture}
    \begin{axis}[%
    width=2.2119in,%
    height=1.8183in,%
    at={(0.758in,0.481in)},%
    scale only axis,%
    xmin=0,%
    xmax=12.0,%
    ymax=0.5,%
    ymin=-8,%
    xmajorgrids,%
    ymajorgrids,%
    ylabel style={yshift=-0.155cm, xshift=0cm}, 
    extra x tick labels={ , ,  , },%
    ylabel={[-]},%
    ytick={0.5,0,-1,-2,-3,-4,-5,-6,-7,-8},%
    yticklabels={\empty,0,\empty,-2,\empty,-4,\empty,-6,\empty,-8},%
    xtick={0,2,...,12},%
    xticklabels={ , , },%
    axis background/.style={fill=white},%
    legend style={at={(0.695,0.95)},anchor=north,legend cell align=left,draw=none,legend 
    columns=-1,align=left,draw=white!15!black}
    ]
    \addplot [color=blue, solid, line width=1.25pt] 
        file{matlabPlots/elevation-linkStress/nominal/s_fR1.txt};%
    \addplot [color=red, solid, line width=1.25pt] 
        file{matlabPlots/elevation-linkStress/nominal/s_fR2.txt};%
    \legend{\footnotesize{$s_{f_{R_1}}$}, \footnotesize{$s_{f_{R_2}}$}};%
    \end{axis}
    \end{tikzpicture}
    }
\end{subfigure}
\\
\vspace{0.025cm}
\hspace{-0.905cm}
\begin{subfigure}{0.425\columnwidth}
    \hspace*{0.095cm}
    \centering
    \scalebox{0.62}{
    \begin{tikzpicture}
    \begin{axis}[%
    width=2.2119in,%
    height=1.8183in,%
    at={(0.758in,0.481in)},%
    scale only axis,%
    xmin=0,%
    xmax=12.0,%
    ymax=0.1,%
    ymin=-1.6,%
    xmajorgrids,%
    ymajorgrids,%
    ylabel style={yshift=-0.255cm, xshift=0cm}, 
    xlabel={Time [\si{\second}]},%
    extra x tick labels={ , ,  , },%
    ylabel={[-]},%
    ytick={0.1,0,-0.2,...,-1.6},%
    yticklabels={0,\empty,-0.2,\empty,-0.4,\empty,-0.6,\empty,-0.8,\empty,1.0,\empty,-1.2,\empty,-1.4,\empty,-1.6},%
    xtick={0,2,...,12},%
    axis background/.style={fill=white},%
    legend style={at={(0.395,0.17)},anchor=north,legend cell align=left,draw=none,legend 
    columns=-1,align=left,draw=white!15!black}
    ]
    \addplot [color=blue, solid, line width=1.25pt] 
        file{matlabPlots/elevation-linkStress/tube/s_delta.txt};%
    \draw [line width=1.25pt, dashed] (rel axis cs: 0, 0.94) -- (rel axis cs: 12, 0.94);
    \legend{\footnotesize{$s_{\Delta}$}};%
    \end{axis}
    \end{tikzpicture}
    }
\end{subfigure}
\hspace*{0.50cm}
\begin{subfigure}{0.425\columnwidth}
    \hspace*{0.095cm}
    \centering
    \scalebox{0.62}{
    \begin{tikzpicture}
    \begin{axis}[%
    width=2.2119in,%
    height=1.8183in,%
    at={(0.758in,0.481in)},%
    scale only axis,%
    xmin=0,%
    xmax=12.0,%
    ymax=0.1,%
    ymin=-8,%
    xmajorgrids,%
    ymajorgrids,%
    ylabel style={yshift=-0.155cm, xshift=0cm}, 
    xlabel={Time [\si{\second}]},%
    extra x tick labels={ , ,  , },%
    ylabel={[-]},%
    ytick={0.5,0,-1,-2,-3,-4,-5,-6,-7,-8},%
    yticklabels={\empty,0,\empty,-2,\empty,-4,\empty,-6,\empty,-8},%
    xtick={0,2,...,12},%
    axis background/.style={fill=white},%
    legend style={at={(0.695,0.95)},anchor=north,legend cell align=left,draw=none,legend 
    columns=-1,align=left,draw=white!15!black}
    ]
    \addplot [color=blue, solid, line width=1.25pt] 
        file{matlabPlots/elevation-linkStress/tube/s_fR1.txt};%
    \addplot [color=red, solid, line width=1.25pt] 
        file{matlabPlots/elevation-linkStress/tube/s_fR2.txt};%
    \legend{\footnotesize{$s_{f_{R_1}}$}, \footnotesize{$s_{f_{R_2}}$}};%
    \end{axis}
    \end{tikzpicture}
    }
\end{subfigure}
    \vspace{-0.5em}
    \caption{Signed constraint residuals for nominal \ac{NMPC} (top) and tube \ac{NMPC} (bottom) under parametric uncertainty.}
    \label{fig:constraint_residual}
\end{figure}

To quantify this, we define signed residuals $s_\Delta(t) \triangleq \cos\!\big(\varphi_2(t)-\varphi_1(t)\big) - \cos(\Delta\varphi_{\min})$ and $s_{f_{R_j}}(t)\triangleq f_{R_j}(t)-f_{R,\max},\ j\in\{1,2\}$, so that $s_\bullet(t) \leq 0$ indicates satisfaction and $s_\bullet(t) = 0$ indicates activation.

Figure~\ref{fig:constraint_residual} compares the resulting residuals obtained with nominal and tube \ac{NMPC} under parametric uncertainty. In the shown run, both controllers satisfy the constraint, but the tube \ac{NMPC} maintains residuals that are consistently more negative, indicating larger safety margins relative to the constraint boundaries. By contrast, nominal \ac{NMPC} tends closer to activation, which is consistent with the absence of proactive tightening when optimizing only for the nominal parameter vector $\mathbf{p}_0$. These results support the intended effect of sensitivity-based tightening: improving robustness of constraint satisfaction with limited impact on tracking performance in low-activation regimes.

\textbf{Monte Carlo.} We run $N_{\mathrm{sim}}=500$ closed-loop trials with $\mathbf{p}$ drawn uniformly from $\mathcal{P}$ (see Table~\ref{tab:parameters}); all other settings are identical between controllers. A trial succeeds if no solver failure occurs and constraints are not violated beyond $\varepsilon_{\mathrm{tol}}=10^{-3}$. We record \ac{RMSE} and signed residuals $s_\Delta$, $s_{f_{R_1}}$, $s_{f_{R_2}}$ per trial.

Over $N_{\mathrm{sim}}=500$ Monte-Carlo runs, the nominal \ac{NMPC} succeeds in $58.0\%$ of the trials, whereas the proposed tube \ac{NMPC} succeeds in $100.0\%$. This gap indicates that sensitivity-based tightening substantially reduces uncertainty-induced failures in this constraint-critical maneuver.

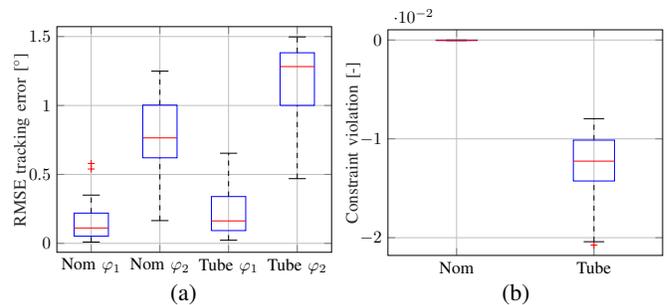
\begin{figure}[tb]
    \centering
\hspace{-0.875cm}
\begin{subfigure}{0.425\columnwidth}
    \centering
    \addtolength{\abovecaptionskip}{-5pt}
    \scalebox{0.65}{
    \begin{tikzpicture}
        \begin{axis}[%
        width=2.2119in,%
        height=1.8183in,%
        at={(0.758in,0.481in)},%
        scale only axis,
        unbounded coords=jump,
        xmin=0.5,
        xmax=4.5,
        xtick={1,2,3,4},
        xticklabels={{Nom $\varphi_1$},{Nom $\varphi_2$},{Tube $\varphi_1$},{Tube $\varphi_2$}},
        ymin=-0.0649014183944121,
        ymax=1.57183913840033,
        ylabel style={yshift=-0.255cm, xshift=0cm, font=\color{white!15!black}},
        ylabel={RMSE tracking error [\si{\degree}]},
        axis background/.style={fill=white},
        xmajorgrids,
        ymajorgrids,
        legend style={legend cell align=left, align=left, draw=white!15!black}
        ]
        \addplot [color=black, dashed, forget plot]
        table[row sep=crcr]{%
        1 0.219311196111804\\
        1 0.350021805505445\\
        };
        
        \addplot [color=black, dashed, forget plot]
        table[row sep=crcr]{%
        2 1.00266622115894\\
        2 1.24919154331333\\
        };
        
        \addplot [color=black, dashed, forget plot]
        table[row sep=crcr]{%
        3 0.339406962153338\\
        3 0.653674908792091\\
        };
        
        \addplot [color=black, dashed, forget plot]
        table[row sep=crcr]{%
        4 1.38238057633725\\
        4 1.49741807947965\\
        };
        
        \addplot [color=black, dashed, forget plot]
        table[row sep=crcr]{%
        1 0.00948564611207813\\
        1 0.0520346105624486\\
        };
        
        \addplot [color=black, dashed, forget plot]
        table[row sep=crcr]{%
        2 0.164882903928303\\
        2 0.620304315091635\\
        };
        
        \addplot [color=black, dashed, forget plot]
        table[row sep=crcr]{%
        3 0.0228770356460649\\
        3 0.0925370270618553\\
        };
        
        \addplot [color=black, dashed, forget plot]
        table[row sep=crcr]{%
        4 0.469237110257047\\
        4 1.00055893334828\\
        };
        
        \addplot [color=black, forget plot]
        table[row sep=crcr]{%
        0.875 0.350021805505445\\
        1.125 0.350021805505445\\
        };
        
        \addplot [color=black, forget plot]
        table[row sep=crcr]{%
        1.875 1.24919154331333\\
        2.125 1.24919154331333\\
        };
        
        \addplot [color=black, forget plot]
        table[row sep=crcr]{%
        2.875 0.653674908792091\\
        3.125 0.653674908792091\\
        };
        
        \addplot [color=black, forget plot]
        table[row sep=crcr]{%
        3.875 1.49741807947965\\
        4.125 1.49741807947965\\
        };
        
        \addplot [color=black, forget plot]
        table[row sep=crcr]{%
        0.875 0.00948564611207813\\
        1.125 0.00948564611207813\\
        };
        
        \addplot [color=black, forget plot]
        table[row sep=crcr]{%
        1.875 0.164882903928303\\
        2.125 0.164882903928303\\
        };
        
        \addplot [color=black, forget plot]
        table[row sep=crcr]{%
        2.875 0.0228770356460649\\
        3.125 0.0228770356460649\\
        };
        
        \addplot [color=black, forget plot]
        table[row sep=crcr]{%
        3.875 0.469237110257047\\
        4.125 0.469237110257047\\
        };
        
        \addplot [color=blue, forget plot]
        table[row sep=crcr]{%
        0.75 0.0520346105624486\\
        0.75 0.219311196111804\\
        1.25 0.219311196111804\\
        1.25 0.0520346105624486\\
        0.75 0.0520346105624486\\
        };
        
        \addplot [color=blue, forget plot]
        table[row sep=crcr]{%
        1.75 0.620304315091635\\
        1.75 1.00266622115894\\
        2.25 1.00266622115894\\
        2.25 0.620304315091635\\
        1.75 0.620304315091635\\
        };
        
        \addplot [color=blue, forget plot]
        table[row sep=crcr]{%
        2.75 0.0925370270618553\\
        2.75 0.339406962153338\\
        3.25 0.339406962153338\\
        3.25 0.0925370270618553\\
        2.75 0.0925370270618553\\
        };
        
        \addplot [color=blue, forget plot]
        table[row sep=crcr]{%
        3.75 1.00055893334828\\
        3.75 1.38238057633725\\
        4.25 1.38238057633725\\
        4.25 1.00055893334828\\
        3.75 1.00055893334828\\
        };
        
        \addplot [color=red, forget plot]
        table[row sep=crcr]{%
        0.75 0.111519707718948\\
        1.25 0.111519707718948\\
        };
        
        \addplot [color=red, forget plot]
        table[row sep=crcr]{%
        1.75 0.764799609620191\\
        2.25 0.764799609620191\\
        };
        
        \addplot [color=red, forget plot]
        table[row sep=crcr]{%
        2.75 0.162723314331165\\
        3.25 0.162723314331165\\
        };
        
        \addplot [color=red, forget plot]
        table[row sep=crcr]{%
        3.75 1.28220187080757\\
        4.25 1.28220187080757\\
        };
        
        \addplot [color=black, draw=none, mark=+, mark options={solid, red}, forget plot]
        table[row sep=crcr]{%
        1 0.539029107899839\\
        1 0.579258633410642\\
        };
        
        \addplot [color=black, draw=none, mark=+, mark options={solid, red}, forget plot]
        table[row sep=crcr]{%
        nan nan\\
        };
        
        \addplot [color=black, draw=none, mark=+, mark options={solid, red}, forget plot]
        table[row sep=crcr]{%
        nan nan\\
        };
        
        \addplot [color=black, draw=none, mark=+, mark options={solid, red}, forget plot]
        table[row sep=crcr]{%
        nan nan\\
        };
    
        \end{axis}
    \end{tikzpicture}
    }
    \caption{\hspace*{-4em}}
    \label{subfig:boxplotRMSE}
\end{subfigure}
\hspace*{0.50cm}
\begin{subfigure}{0.425\columnwidth}
    \centering
    \addtolength{\abovecaptionskip}{-5pt}
    \scalebox{0.65}{
    \begin{tikzpicture}
        \begin{axis}[
        width=2.2119in,%
        height=1.8183in,%
        at={(0.758in,0.481in)},%
        scale only axis,
        unbounded coords=jump,
        xmin=0.5,
        xmax=2.5,
        xtick={1,2},
        xticklabels={{Nom},{Tube}},
        ymin=-0.0217940779471488,
        ymax=0.00103224422766156,
        ylabel={Constraint violation [-]},
        ylabel style={yshift=-0.255cm, xshift=0cm,font=\color{white!15!black}},
        axis background/.style={fill=white},
        xmajorgrids,
        ymajorgrids,
        legend style={
        legend cell align=left,
        align=left,
        draw=white!15!black
        }
        ]
        
        \addplot [color=black, dashed, forget plot]
          table[row sep=crcr]{%
        1	-1.27777587056732e-05\\
        1	-5.31587119345645e-06\\
        };
        \addplot [color=black, dashed, forget plot]
          table[row sep=crcr]{%
        2	-0.0101316515205819\\
        2	-0.00796139401596085\\
        };
        \addplot [color=black, dashed, forget plot]
          table[row sep=crcr]{%
        1	-5.66260131144025e-05\\
        1	-4.4272836985576e-05\\
        };
        \addplot [color=black, dashed, forget plot]
          table[row sep=crcr]{%
        2	-0.0204068350382173\\
        2	-0.0142638769108491\\
        };
        \addplot [color=black, forget plot]
          table[row sep=crcr]{%
        0.925	-5.31587119345645e-06\\
        1.075	-5.31587119345645e-06\\
        };
        \addplot [color=black, forget plot]
          table[row sep=crcr]{%
        1.925	-0.00796139401596085\\
        2.075	-0.00796139401596085\\
        };
        \addplot [color=black, forget plot]
          table[row sep=crcr]{%
        0.925	-5.66260131144025e-05\\
        1.075	-5.66260131144025e-05\\
        };
        \addplot [color=black, forget plot]
          table[row sep=crcr]{%
        1.925	-0.0204068350382173\\
        2.075	-0.0204068350382173\\
        };
        \addplot [color=blue, forget plot]
          table[row sep=crcr]{%
        0.85	-4.4272836985576e-05\\
        0.85	-1.27777587056732e-05\\
        1.15	-1.27777587056732e-05\\
        1.15	-4.4272836985576e-05\\
        0.85	-4.4272836985576e-05\\
        };
        \addplot [color=blue, forget plot]
          table[row sep=crcr]{%
        1.85	-0.0142638769108491\\
        1.85	-0.0101316515205819\\
        2.15	-0.0101316515205819\\
        2.15	-0.0142638769108491\\
        1.85	-0.0142638769108491\\
        };
        \addplot [color=red, forget plot]
          table[row sep=crcr]{%
        0.85	-3.32031662302334e-05\\
        1.15	-3.32031662302334e-05\\
        };
        \addplot [color=red, forget plot]
          table[row sep=crcr]{%
        1.85	-0.0122552811786465\\
        2.15	-0.0122552811786465\\
        };
        \addplot [color=black, draw=none, mark=+, mark options={solid, red}, forget plot]
          table[row sep=crcr]{%
        nan	nan\\
        };
        \addplot [color=black, draw=none, mark=+, mark options={solid, red}, forget plot]
          table[row sep=crcr]{%
        2	-0.0207565178482938\\
        };
        \end{axis}
    \end{tikzpicture}
    }
    \caption{\hspace*{-4em}}
    \label{subfig:boxplotSignedConstraint}
\end{subfigure}
    \vspace{-0.25em}
    \caption{Boxplots for the Monte-Carlo campaign, comparing nominal \ac{NMPC} and sensitivity-based tube \ac{NMPC}.}
    \vspace{-0.5em}
\end{figure}

Figure~\ref{subfig:boxplotRMSE} reports boxplots of the elevation-angle tracking \ac{RMSE} for $\varphi_1$ and $\varphi_2$. In this experiment, tube \ac{NMPC} does not systematically improve tracking accuracy relative to nominal \ac{NMPC}: the medians and spreads are comparable, and in some cases tube \ac{NMPC} exhibits a larger dispersion. This behavior is consistent with the expected performance--robustness trade-off introduced by constraint tightening, which may prioritize safety margins in constraint-relevant regions at the expense of nominal tracking optimality.

Figure~\ref{subfig:boxplotSignedConstraint} reports boxplots of these residuals. Under nominal \ac{NMPC}, the distribution is concentrated near the activation threshold (approximately a line at zero), indicating operation close to constraint activation across trials. In contrast, tube \ac{NMPC} shifts the distribution to negative values, meaning that the constraints are satisfied with a nonzero safety margin across uncertainty realizations, including worst-case trials. 

These results confirm that sensitivity-based tightening substantially improves robustness while preserving tracking performance, at the cost of increased solve time discussed next.

\textbf{Computation performance.} The augmented prediction model has approximately $84(N+1)+4N = 2724$ decision variables for $N=30$, compared to $12(N+1)+4N = 492$ for nominal \ac{NMPC}. 
In our MATLAB prototype using \textsc{MATMPC} with \textsc{qpOASES}, the mean solve time over the Monte-Carlo campaign is $6.7\pm1.5~\si{\milli\second}$ for nominal \ac{NMPC} and $21.7\pm5.3~\si{\milli\second}$ for tube \ac{NMPC}, yielding an overhead factor of approximately $3.2\times$ for a $\sim\!7\times$ increase in decision variables. 
Both solve times reflect MATLAB interpretation overhead rather than the method's intrinsic cost; a code-generated embedded implementation (e.g., \textsc{acados}\footnote{\url{https://docs.acados.org/}}) is expected to recover real-time feasibility at \SI{100}{\hertz}, as demonstrated for comparable NLP sizes in~\cite{Nyboe2025ICRA}.



\section{Conclusions}
\label{sec:conclusions}

This paper presented a sensitivity-based tube \ac{NMPC} framework for cooperative aerial chains under parametric uncertainty. By propagating parametric sensitivities and applying constraint tightening, the proposed approach improves constraint robustness while preserving real-time feasibility. Monte-Carlo simulations demonstrate improved success rate and constraint satisfaction compared to nominal \ac{NMPC}. Future work will extend the approach to three-dimensional systems, investigate tighter robustness bounds (e.g., via higher-order sensitivity terms or interval arithmetic), and consider cable links, where unilateral tension constraints and variable effective length introduce additional modeling and robustness challenges.



\bibliographystyle{IEEEtran}
\bibliography{references}

\end{document}